\documentclass[sigconf]{acmart}
\AtBeginDocument{%
  }

\setcopyright{acmlicensed}

\copyrightyear{2025}
\acmYear{2025}
\setcopyright{acmlicensed}\acmConference[KDD '25]{Proceedings of the 31st ACM SIGKDD Conference on Knowledge Discovery and Data Mining V.2}{August 3--7, 2025}{Toronto, ON, Canada}
\acmBooktitle{Proceedings of the 31st ACM SIGKDD Conference on Knowledge Discovery and Data Mining V.2 (KDD '25), August 3--7, 2025, Toronto, ON, Canada}
\acmDOI{10.1145/3711896.3737379}
\acmISBN{979-8-4007-1454-2/2025/08}

\usepackage[capitalize]{cleveref}

\usepackage{xspace}
\usepackage{graphicx}
\usepackage{caption}
\usepackage{subfigure}
\usepackage{booktabs}
\usepackage{bm}
\usepackage{xcolor}
\usepackage{float}
\usepackage{multirow}
\usepackage{longtable}
\usepackage{framed}
\usepackage{enumitem}
\usepackage{amsthm,amsmath}
\usepackage{pythonhighlight}
\usepackage{appendix}
\usepackage[utf8]{inputenc}
\usepackage{bbding}
\usepackage{tabularx}
\usepackage{dsfont}

\DeclareMathAlphabet{\mathpzc}{OT1}{pzc}{m}{it}

\usepackage{mathtools}
\usepackage{multirow}
\usepackage{algorithm}
\usepackage[noend]{algpseudocode}
\usepackage{color,soul}

\makeatletter
\DeclareRobustCommand\onedot{\futurelet\@let@token\@onedot}
\def\@onedot{\ifx\@let@token.\else.\null\fi\xspace}
\def\eg{\emph{e.g}\onedot} 
\def\ie{\emph{i.e}\onedot} 
 
\def\etc{\emph{etc}\onedot}

\makeatother

\definecolor{blue_}{RGB}{76, 114, 176}
\definecolor{orange_}{RGB}{221, 132, 82}
\definecolor{upload}{RGB}{47, 85, 151}
\definecolor{download}{RGB}{241, 13, 208}
\definecolor{red_}{RGB}{255, 0, 0}
\definecolor{gray_}{RGB}{127, 127, 127}
\definecolor{green_}{RGB}{1, 128, 0}
\definecolor{sjtured_}{RGB}{192, 0, 0}
\definecolor{sjtugreen_}{RGB}{84, 130, 53}
\definecolor{hist_red}{RGB}{194, 82, 83}
\definecolor{hist_blue}{RGB}{83, 110, 174}

\usepackage{pifont}
\usepackage{makecell}
\usepackage{colortbl}
\definecolor{grayline}{gray}{0.9}

\crefname{section}{Sec.}{Secs.}
\Crefname{section}{Section}{Sections}
\Crefname{table}{Table}{Tables}
\crefname{table}{Tab.}{Tabs.}
\crefname{figure}{Fig.}{Figs.}
\crefname{equation}{Eq.}{Eqs.}

\usepackage[utf8]{inputenc} 
\usepackage[T1]{fontenc}    
\usepackage{url}            
\usepackage{booktabs}       
\usepackage{amsfonts}       
\usepackage{nicefrac}       
\usepackage{microtype}      

\usepackage{wrapfig}
\usepackage{balance}

\def\ours{\texttt{HtFLlib}\xspace}

\begin{document}

\title{\ours: A Comprehensive Heterogeneous Federated Learning Library and Benchmark}

\author{Jianqing Zhang}
\authornote{Jianqing Zhang is also affiliated with the Institute for AI Industry Research, Tsinghua University, Beijing, China.}
\affiliation{%
  \institution{Shanghai Jiao Tong University}
  \state{Shanghai}
  \country{China}
}
\email{tsingz@sjtu.edu.cn}

\author{Xinghao Wu}
\affiliation{%
  \institution{Beihang University}
  \state{Beijing}
  \country{China}
}
\email{wuxinghao@buaa.edu.cn}

\author{Yanbing Zhou}
\affiliation{%
  \institution{Chongqing University}
  \state{Chongqing}
  \country{China}
}
\email{202124021011@cqu.edu.cn}

\author{Xiaoting Sun}
\affiliation{%
  \institution{Tongji University}
  \state{Shanghai}
  \country{China}
}
\email{tsxt@tongji.edu.cn}

\author{Qiqi Cai}
\affiliation{%
  \institution{Shanghai Jiao Tong University}
  \city{Shanghai}
  \country{China}
}
\email{cai_qiqi@sjtu.edu.cn}

\author{Yang Liu}
\authornote{Yang Liu is a corresponding author and is also affiliated with the Shanghai Artificial Intelligence Laboratory, Shanghai, China.}
\affiliation{%
  \institution{Hong Kong Polytechnic University}
  \state{Hong Kong}
  \country{China}
}
\email{yang-veronica.liu@polyu.edu.hk}

\author{Yang Hua}

\affiliation{%
  \institution{The Queen's University of Belfast}
  \state{Belfast}
  \country{UK}
}
\email{y.hua@qub.ac.uk}

\author{Zhenzhe Zheng}
\affiliation{%
  \institution{Shanghai Jiao Tong University}
  \state{Shanghai}
  \country{China}
}
\email{zzheng@cs.sjtu.edu.cn}

\author{Jian Cao}
\authornote{Jian Cao is a corresponding author and is also affiliated with the Shanghai Key Laboratory of Trusted Data Circulation and Governance in Web3, Shanghai, China.}
\affiliation{%
  \institution{Shanghai Jiao Tong University}
  \state{Shanghai}
  \country{China}
}
\email{cao-jian@sjtu.edu.cn}

\author{Qiang Yang}
\affiliation{%
  \institution{Hong Kong Polytechnic University}
  \state{Hong Kong}
  \country{China}
}
\email{profqiang.yang@polyu.edu.hk}

\renewcommand{\shortauthors}{Jianqing Zhang et al.}


\begin{abstract}
As AI evolves, collaboration among heterogeneous models helps overcome data scarcity by enabling knowledge transfer across institutions and devices. 
Traditional Federated Learning (FL) only supports homogeneous models, limiting collaboration among clients with heterogeneous model architectures. To address this, Heterogeneous Federated Learning (HtFL) methods are developed to enable collaboration across diverse heterogeneous models while tackling the data heterogeneity issue at the same time. 
However, a comprehensive benchmark for standardized evaluation and analysis of the rapidly growing HtFL methods is lacking. Firstly, the highly varied datasets, model heterogeneity scenarios, and different method implementations become hurdles to making easy and fair comparisons among HtFL methods. Secondly, the effectiveness and robustness of HtFL methods are under-explored in various scenarios, such as the medical domain and sensor signal modality. 
To fill this gap, we introduce the first \textbf{Heterogeneous Federated Learning Library (\ours)}, an easy-to-use and extensible framework that integrates multiple datasets and model heterogeneity scenarios, offering a robust benchmark for research and practical applications. 
Specifically, \ours integrates (1) 12 datasets spanning various domains, modalities, and data heterogeneity scenarios; (2) 40 model architectures, ranging from small to large, across three modalities; (3) a modularized and easy-to-extend HtFL codebase with implementations of 10 representative HtFL methods; and (4) systematic evaluations in terms of accuracy, convergence, computation costs, and communication costs. 
We emphasize the advantages and potential of state-of-the-art HtFL methods and hope that \ours will catalyze advancing HtFL research and enable its broader applications. 
The code is released at \url{https://github.com/TsingZ0/HtFLlib}. 
\end{abstract}

\ccsdesc[500]{Computing methodologies~Distributed artificial intelligence}
\ccsdesc[500]{Security and privacy~Privacy protections}

\keywords{Heterogeneous Federated Learning, Benchmark, Model Heterogeneity, Data Heterogeneity}
\begin{teaserfigure}
    \includegraphics[width=\textwidth]{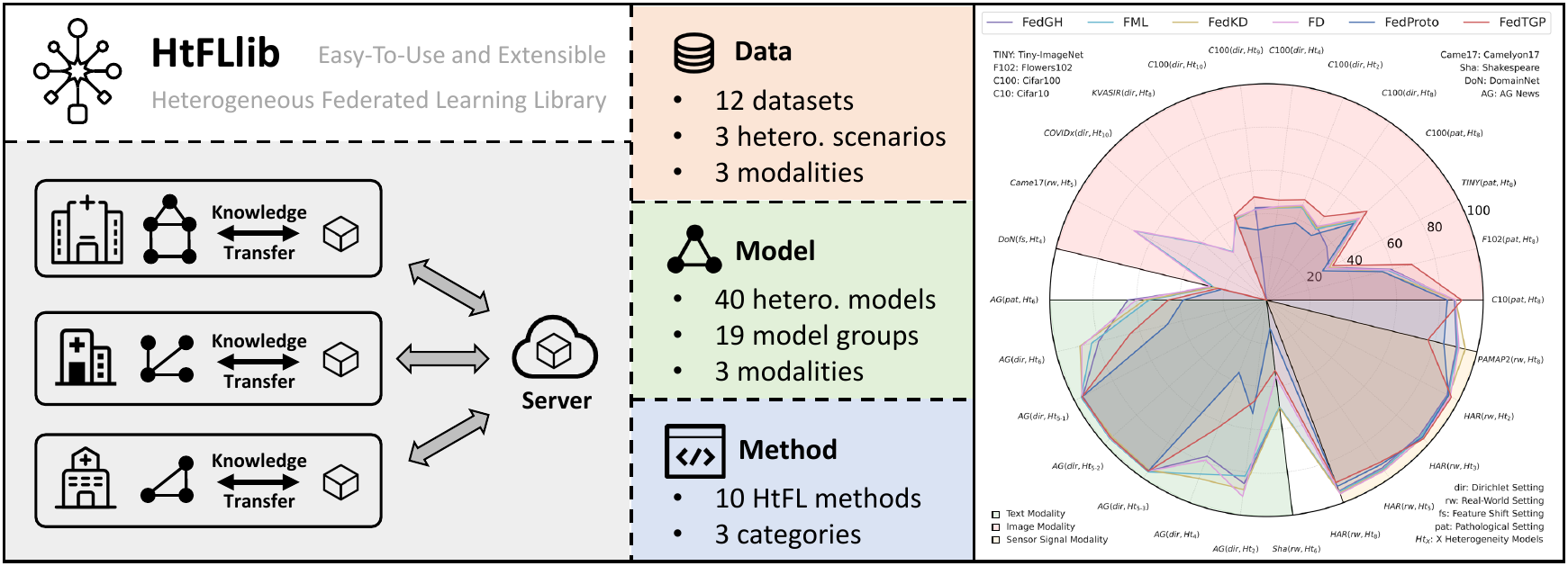}
    \caption{Overview of \ours along with experimental results for representative HtFL methods across various heterogeneous model groups, modalities, and data scenarios. Left: Lightweight knowledge carriers \includegraphics[width=0.012\textwidth]{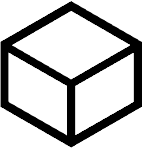} are exchanged between the server and clients for \textit{knowledge transfer}, as sharing entire models is infeasible. Right: Results indicate that methods like FD consistently perform well, while others like FedTGP demonstrate superiority primarily in image tasks. \textit{Best viewed zoomed in}.}
    \label{fig:teaser}
\end{teaserfigure}


\maketitle

\setlength{\abovecaptionskip}{4pt}

\section{Introduction}

As AI advances, diverse institutions develop heterogeneous models tailored to specific tasks \cite{hao2024one, gao2024federated} but face data scarcity during training \cite{nguyen2021federated, wei2025trustworthy}. Collaboration among these models enables knowledge transfer, overcoming data access limitations while leveraging shared expertise \cite{zhang2024fedtgp, yi2024federated}. 
Federated Learning (FL) is a widely recognized privacy-preserving collaborative learning technique that enables knowledge transfer among participating clients \cite{kairouz2019advances}. Notably, traditional FL is limited to supporting collaboration among homogeneous models, requiring all clients to use identical architectures \cite{yang2019federated}. However, clients often develop specialized model architectures tailored to their unique requirements \cite{huang2020unet, ma2024segment}. Additionally, sharing effort-intensive locally trained models can compromise intellectual property (IP) \cite{zhang2024fedtgp}. The requirement to use homogeneous models and share entire local models reduces participants' willingness to engage in collaborations \cite{zhang2024upload}. 

Heterogeneous Federated Learning (HtFL) has emerged as a rapidly growing research area that allows participants to collaborate using their heterogeneous models \cite{tan2022fedproto, zhang2024upload, yi2023fedgh}, broadening the scope of traditional FL and fostering wider participation. 
In a typical HtFL framework, participating clients collaborate to improve their heterogeneous models with local private data by communicating and aggregating lightweight knowledge carriers in a federated manner, as sharing entire models is infeasible \cite{ye2023heterogeneous, yi2024federated, zhang2024fedtgp}. 

In the literature, there is currently \textbf{\textit{no}} benchmark for HtFL that offers unified and standard scenarios to evaluate HtFL methods in various domains and aspects. 
To be specific: 
\begin{itemize}[left=0pt, labelsep=0.6em]
    \item \textbf{Non-unified datasets, model heterogeneity, and implementations for HtFL.} Due to the lack of a unified HtFL benchmark, researchers created custom experimental setups with varying data and model heterogeneity. For example, \cite{tan2022fedproto} uses MNIST \cite{lecun1998gradient}, FEMNIST \cite{caldas2018leaf}, and Cifar10 \cite{krizhevsky2009learning} with specific client data partition, while \cite{yi2024federated} applies Dirichlet distribution-based partition for Cifar10/100 \cite{krizhevsky2009learning}. Besides, \cite{yi2023fedgh} focuses on heterogeneous CNNs, and \cite{zhang2024fedtgp} and \cite{zhang2024upload} explore collaboration between ResNets \cite{he2016deep} and ViTs \cite{dosovitskiy2020image}. Moreover, the choice of optimizers, batch size, learning rate, \etc., significantly impacts the results \cite{zhou2020towards}. 
    \item \textbf{Under-explored applicability of HtFL across diverse scenarios.} Current HtFL methods primarily evaluate effectiveness on common image datasets in simulated partitions \cite{tan2022fedproto, zhang2024fedtgp, zhang2024upload, yi2023fedgh, yi2024federated}, overlooking other modalities (such as text and sensor signals) in real-world settings and specialized domains like medicine, where collaboration among heterogeneous models is practical and valuable \cite{castiglioni2021ai, cao2022ai}. However, it remains unclear whether existing HtFL methods perform consistently across diverse scenarios. 
\end{itemize}
To accelerate progress in HtFL, we introduce the \textbf{\textit{first}} HtFL benchmark \textbf{Heterogeneous Federated Learning Library (\ours)}, as illustrated in \cref{fig:teaser}. 
Specifically, our contributions are:
\begin{itemize}[left=0pt, labelsep=0.6em]
    \item We offer 3 benchmark families for image, text, and sensor signal, featuring 40 heterogeneous model architectures and 12 datasets covering label skew, feature shift, and real-world scenarios, each with unified data and model heterogeneity settings. 
    \item We open-source an easily extensible HtFL codebase featuring 10 representative methods, with unified interfaces and modular components, so that only a small portion of essential modules need to be modified when adding new methods. 
    \item We conduct systematic evaluations of HtFL methods, providing reproducible results on key aspects such as accuracy, convergence, computation, and communication costs. Additionally, we highlight the advantages of HtFL methods and offer insights for future research in the field. 
\end{itemize}

\section{Background}

\subsection{Existing FL Benchmarks}

In the past, numerous benchmarks have been proposed to assess the performance of FL methods \cite{liu2021fate, he2020fedml, federatedscope, beutel2020flower, wu2022motley, hu2022oarf, chai2020fedeval, zeng2023fedlab, li2022federated, zhang2025pfllib}.
Most previous benchmarks primarily focus on data heterogeneity using homogeneous client models, neglecting model heterogeneity. However, scenarios involving heterogeneous models are more practical, as the research and application of AI models have been ongoing for years, and many organizations have already developed their specific model architectures for their needs \cite{sarker2021deep, alom2019state}. 
\ours addresses this gap by incorporating up to 40 heterogeneous models across experiments. Specifically, it supports 19 heterogeneous model groups, each assigned to clients to implement the model heterogeneity scenario for each experiment. In this way, \ours can advance the study of HtFL, enabling more flexible and effective collaborative learning. 

\subsection{Representative HtFL Methods}
\label{sec:methods}

We categorize existing HtFL methods into three main categories: (1) partial parameter sharing, (2) mutual distillation, and (3) prototype sharing. For each category, we select several representative methods for our benchmark. 
Within the three categories of HtFL, our \ours includes 10 state-of-the-art methods, as described below.

(1) \textit{partial parameter sharing}: These methods allow the main parts of clients' models to remain heterogeneous while assuming the remaining lightweight components (\eg, classifier heads) are homogeneous for knowledge transfer. For example, LG-FedAvg \cite{liang2020think}, FedGen \cite{zhu2021data}, and FedGH \cite{yi2023fedgh} decouple each client model into a heterogeneous feature extractor and a homogeneous classifier head. In LG-FedAvg, clients send the parameters of their classifier head to the server for aggregation, whereas FedGH trains a global classifier head on the server using uploaded class-wise feature representations (\ie, prototypes).
In contrast, FedGen trains a small generator on the server to produce general features for aligning clients' classifiers in the feature space. 

(2) \textit{mutual distillation}: Methods such as FML \cite{shen2020federated}, FedKD \cite{wu2022communication}, and FedMRL \cite{yi2024federated} simultaneously train and share a small auxiliary model using mutual distillation \cite{zhang2018deep}. FML guides the training of both the auxiliary model and heterogeneous client models by sharing output logits. Compared to FML, FedKD additionally aligns intermediate feature vectors, while FedMRL combines the features extracted by the auxiliary and local models during inference. 

(3) \textit{prototype sharing}: These methods transfer lightweight class-wise prototypes as global knowledge. Local prototypes are collected from each client, aggregated on the server to create global prototypes, and then used to guide local training on clients. The key differences among these methods lie in the dimensionality of the prototypes. For instance, FD \cite{jeong2018communication} applies prototype guidance in the logit space, while FedProto \cite{tan2022fedproto} and FedTGP \cite{zhang2024fedtgp} use the intermediate feature space to generate and refine prototypes. FedTGP further adaptively enhances the discriminability among global prototypes to improve their quality. FedKTL \cite{zhang2024upload} goes a step further by using a server-side pre-trained large generator to generate images corresponding to prototypes, enriching local training with image-prototype pairs, but FedKTL only applies to image tasks.

\subsection{Problem Statement of HtFL}

In HtFL, \(N\) clients participate in collaborative learning, each bringing their respective heterogeneous models with parameters \(\theta_1, \ldots, \theta_N\) and heterogeneous training data \(\mathcal{D}_1, \ldots, \mathcal{D}_N\). These clients learn from each other through a shared global knowledge carrier \(\mathcal{S}_g\), which is obtained by aggregating the clients' shared local knowledge \(\mathcal{S}_i\) on a central server. Formally, the objective is to iteratively optimize the following formula in a federated manner:

\vspace{-5pt}
\begin{equation}
    \min_{\theta_1, \ldots, \theta_N} \ \sum_{i=1}^N \frac{k_i}{k} \mathcal{L}_i(\theta_i; \mathcal{D}_i, \mathcal{S}_g),
\end{equation}

where \(k_i\) is the size of the training set \(\mathcal{D}_i\), \(k = \sum_{i=1}^N k_i\) and \(\mathcal{L}_i\) is the local training objective. Typically, \(\mathcal{S}_g = \frac{k_i}{k} \mathcal{S}_i\). 
The definitions of \(\mathcal{L}_i\), \(\mathcal{S}_i\), and \(\mathcal{S}_g\) vary across different HtFL methods. In partial parameter sharing, \(\mathcal{S}_i\) and \(\mathcal{S}_g\) represent the local and global partial model parameters; in mutual distillation, they refer to the local and global tiny auxiliary models; and in prototype sharing, they denote the local and global prototypes.

\section{Setups and Assets in \ours}

We first introduce the necessary basic setups for all experiments here. More details are provided in the Appendix. 

\subsection{Basic Setups}
\subsubsection{Data heterogeneity scenarios}

\ours includes comprehensive data heterogeneity scenarios, categorized into three settings:

\begin{enumerate}[left=0pt, labelsep=0.6em]
    \item \textbf{Label Skew Setting}: In this scenario, different clients possess data with varying numbers of labels \cite{zhang2022fedala}. This is further divided into two sub-settings:
    \begin{enumerate}[left=0pt, labelsep=0.6em]
        \item \textbf{Pathological Setting}: Each client holds only a subset of the available labels across all clients \cite{mcmahan2017communication}.
        \item \textbf{Dirichlet Setting}: We allocate data of class \(y\) to each client using a client-specific ratio \(q^y\), sampled from a Dirichlet distribution with a control parameter \(\alpha\), leading to a more realistic class imbalance \cite{lin2020ensemble}. By default, we set \(\alpha = 0.1\).
    \end{enumerate}

    \item \textbf{Feature Shift Setting}: Here, clients have an identical number of labels but differ in the features of their data, such as the distinction between sketch images and painting images.

    \item \textbf{Real-World Setting}: In this scenario, the data on each client is naturally collected by an individual user or sensor, representing a real-world data distribution \cite{zhang2025pfllib}.
\end{enumerate}

\subsubsection{Model heterogeneity scenarios}
\label{sec:hete_scene}

In \ours, we adopt the notation HtFE$^{dom}$$_X$, following the convention established in \cite{zhang2024upload}. Here, HtFE$^{dom}$$_X$ represents a group of heterogeneous feature extractors, where \(dom\) indicates the specific domain (\eg, \(img\), \(txt\), and \(sen\) for image, text, and sensor signal, respectively), and \(X\) denotes the degree of model heterogeneity (positively correlated), while the classifier heads remain homogeneous across clients. 
Within each group, such as HtFE$^{dom}$$_X$, the \((i \mod X)\)-th model in the group is assigned to the client \(i\). 
Additionally, we introduce notations HtC$^{dom}$$_X$ and HtM$^{dom}$$_X$ to represent the group of heterogeneous classifiers and fully heterogeneous models, respectively. 
To meet the common requirement of identical feature dimensions (\(K\)) for methods like FedGH, FedKD, FedProto, and FedTGP, we add an average pooling layer \cite{szegedy2015going} before the classifier heads. By default, we set \(K = 512\) for all models to ensure compatibility and consistency across experiments.

\subsection{Assets}

\subsubsection{Baselines} 

Few existing HtFL methods enable knowledge transfer among private clients and support client-specific heterogeneous model architectures. We categorize these methods into three types: (1) \textit{partial parameter sharing}: LG-FedAvg \cite{liang2020think}, FedGen \cite{zhu2021data}, and FedGH \cite{yi2023fedgh}, (2) \textit{mutual distillation}: FML \cite{shen2020federated}, FedKD \cite{wu2022communication}, and FedMRL \cite{yi2024federated}, and (3) \textit{prototype sharing}: FD \cite{jeong2018communication}, FedProto \cite{tan2022fedproto}, FedTGP \cite{zhang2024fedtgp}, and FedKTL \cite{zhang2024upload}. Refer to \cref{sec:methods} for their details. 

\begin{table*}[h]
  \centering
  \caption{Test accuracy (\%) on four datasets under both pathological and practical label skew settings using HtFE$^{img}$$_8$.}
  \resizebox{!}{!}{
    \begin{tabular}{l|*{4}{c}|*{4}{c}}
    \toprule
    Settings & \multicolumn{4}{c|}{Pathological Setting} & \multicolumn{4}{c}{Dirichlet Setting} \\
    \midrule
    Datasets & Cifar10 & Cifar100 & Flowers102 & Tiny-ImageNet & Cifar10 & Cifar100 & Flowers102 & Tiny-ImageNet \\
    \midrule
    LG-FedAvg & \ul{86.82$\pm$0.26} & 57.01$\pm$0.66 & 58.88$\pm$0.28 & 32.04$\pm$0.17 & \ul{84.55$\pm$0.51} & 40.65$\pm$0.07 & 45.93$\pm$0.48 & \ul{24.06$\pm$0.10} \\
    FedGen & 82.83$\pm$0.65 & \ul{58.26$\pm$0.36} & \ul{59.90$\pm$0.15} & 29.80$\pm$1.11 & 82.55$\pm$0.49 & 38.73$\pm$0.14 & 45.30$\pm$0.17 & 19.60$\pm$0.08 \\
    FedGH & 86.59$\pm$0.23 & 57.19$\pm$0.20 & 59.27$\pm$0.33 & \ul{32.55$\pm$0.37} & 84.43$\pm$0.31 & \ul{40.99$\pm$0.51} & \ul{46.13$\pm$0.17} & 24.01$\pm$0.11 \\
    \midrule
    FML & 87.06$\pm$0.24 & 55.15$\pm$0.14 & 57.79$\pm$0.31 & 31.38$\pm$0.15 & 85.88$\pm$0.08 & 39.86$\pm$0.25 & 46.08$\pm$0.53 & 24.25$\pm$0.14 \\
    FedKD & 87.32$\pm$0.31 & 56.56$\pm$0.27 & 54.82$\pm$0.35 & 32.64$\pm$0.36 & \ul{86.45$\pm$0.10} & 40.56$\pm$0.31 & 48.52$\pm$0.28 & 25.51$\pm$0.35 \\
    FedMRL & \ul{87.80$\pm$0.30} & \ul{59.80$\pm$0.50} & \ul{60.90$\pm$0.80} & \ul{33.20$\pm$0.40} & 86.20$\pm$0.40 & \ul{41.20$\pm$0.50} & \ul{48.56$\pm$0.23} & \ul{25.83$\pm$0.31} \\
    \midrule
    FD & 87.24$\pm$0.06 & 56.99$\pm$0.27 & 58.51$\pm$0.34 & 31.49$\pm$0.38 & 86.01$\pm$0.31 & 41.54$\pm$0.08 & 49.13$\pm$0.85 & 24.87$\pm$0.31 \\
    FedProto & 83.39$\pm$0.15 & 53.59$\pm$0.29 & 55.13$\pm$0.17 & 29.28$\pm$0.36 & 82.07$\pm$1.64 & 36.34$\pm$0.28 & 41.21$\pm$0.22 & 19.01$\pm$0.10 \\
    FedTGP & \textbf{\ul{90.02$\pm$0.30}} & 61.86$\pm$0.30 & \textbf{\ul{68.98$\pm$0.43}} & 34.56$\pm$0.27 & \textbf{\ul{88.15$\pm$0.43}} & \textbf{\ul{46.94$\pm$0.12}} & \textbf{\ul{53.68$\pm$0.31}} & 27.37$\pm$0.12\\
    FedKTL & 88.43$\pm$0.13 & \textbf{\ul{62.01$\pm$0.28}} & 64.72$\pm$0.62 & \textbf{\ul{34.74$\pm$0.17}} & 87.63$\pm$0.07 & 46.94$\pm$0.23 & 53.16$\pm$0.08 & \textbf{\ul{28.17$\pm$0.19}} \\
    \bottomrule
    \end{tabular}}
    \label{tab:datasets_acc}
    \vspace{-5pt}
\end{table*}

\subsubsection{Datasets}

In \ours, we provide 12 datasets across three modalities and three data heterogeneity scenarios. Specifically, we list all 12 datasets as follows:
\begin{enumerate}[left=0pt, labelsep=0.6em]
    \item \textbf{Cifar10} \cite{krizhevsky2009learning}: \textit{Modality}: image, \textit{Scenario}: label skew, \textit{Description}: 60K common images across 10 classes.
    \item \textbf{Cifar100} \cite{krizhevsky2009learning}: \textit{Modality}: image, \textit{Scenario}: label skew, \textit{Description}: 60K common images across 100 classes.
    \item \textbf{Flowers102} \cite{nilsback2008automated}: \textit{Modality}: image, \textit{Scenario}: label skew, \textit{Description}: 8K flower images across 102 classes.
    \item \textbf{Tiny-ImageNet} \cite{chrabaszcz2017downsampled}: \textit{Modality}: image, \textit{Scenario}: label skew, \textit{Description}: 100K common images across 200 classes.
    \item \textbf{KVASIR} \cite{pogorelov2017kvasir}: \textit{Modality}: image, \textit{Scenario}: label skew, \textit{Description}: 1K colonoscopy medical images (\eg, esophagitis, polyps, etc.) across 8 classes.
    \item \textbf{COVIDx} \cite{wang2020covid}: \textit{Modality}: image, \textit{Scenario}: label skew, \textit{Description}: 38K chest X-ray images across 2 classes.
    \item \textbf{DomainNet} \cite{peng2019moment}: \textit{Modality}: image, \textit{Scenario}: feature shift, \textit{Description}: 600K images across 6 domains and 345 classes.
    \item \textbf{Camelyon17} \cite{koh2021wilds}: \textit{Modality}: image, \textit{Scenario}: real-world, \textit{Description}: 422K histological lymph node section images across 2 classes collected from 5 hospitals.
    \item \textbf{AG News} \cite{zhang2015character}: \textit{Modality}: text, \textit{Scenario}: label skew, \textit{Description}: 127K articles across 4 classes.
    \item \textbf{Shakespeare} \cite{zhang2015character}: \textit{Modality}: text, \textit{Scenario}: real-world, \textit{Description}: a refined version with 73K lines collected from 118 speaking roles to predict the next character.
    \item \textbf{HAR} \cite{anguita2012human}: \textit{Modality}: sensor signal, \textit{Scenario}: real-world, \textit{Description}: 10K signal across 6 physical activities collected from 30 smartphones with accelerometers and gyroscopes.
    \item \textbf{PAMAP2} \cite{reiss2012introducing}: \textit{Modality}: sensor signal, \textit{Scenario}: real-world, \textit{Description}: 15K signal across 18 physical activities collected from 9 subjects wearing inertial measurement units and a heart rate monitor.
\end{enumerate}
These datasets vary significantly in field, data volume, and the number of classes, showcasing the comprehensive and versatile nature of \ours. While we include datasets from all three modalities, we focus more on image data, especially the label skew setting, as image tasks are the most commonly used tasks in the field \cite{zhang2024fedtgp, yi2024federated, tan2022fedproto, zhu2021data}.

\subsubsection{Heterogeneous model architectures}
Our principle of selecting model architectures is \textit{widely used, with official implementations, various architectures, and diverse capabilities}. After a careful survey, we include 40 heterogeneous model architectures in \ours, organized into 19 distinct groups. Each group is assigned to a specific experiment, as outlined in \cref{sec:hete_scene}, where \(X\) represents the degree of model heterogeneity (positively correlated) for HtFE/HtM/HtC$^{dom}$$_X$. Below are the details of all 19 model groups:
\begin{enumerate}[left=0pt, labelsep=0.6em]
    \item \textbf{HtFE$^{img}$$_2$}: 4-layer CNN \cite{mcmahan2017communication} and ResNet18 \cite{he2016deep}. 
    \item \textbf{HtFE$^{img}$$_3$}: ResNet10 \cite{zhong2017deep}, ResNet18, and ResNet34 \cite{he2016deep}.
    \item \textbf{HtFE$^{img}$$_4$}: 4-layer CNN, GoogleNet \cite{szegedy2015going}, MobileNetv2 \cite{sandler2018mobilenetv2}, and ResNet18. 
    \item \textbf{HtFE$^{img}$$_5$}: GoogleNet, MobileNetv2, ResNet18, ResNet34, and ResNet50 \cite{he2016deep}. 
    \item \textbf{HtFE$^{img}$$_8$}: 4-layer CNN, GoogleNet, MobileNetv2, ResNet18, ResNet34, ResNet50, ResNet101, and ResNet152 \cite{he2016deep}. 
    \item \textbf{HtFE$^{img}$$_9$}: ResNet4, ResNet6, and ResNet8 \cite{zhong2017deep}, ResNet10, ResNet18, ResNet34, ResNet50, ResNet101, and ResNet152. 
    \item \textbf{Res34-HtC$^{img}$$_4$}: ResNet34 with 4 types of heads \cite{zhang2024fedtgp}. 
    \item \textbf{HtFE$^{img}$$_8$-HtC$^{img}$$_4$}: HtFE$^{img}$$_8$ with 4 types of heads \cite{zhang2024fedtgp}. 
    \item \textbf{HtM$^{img}$$_{10}$}: HtFE$^{img}$$_8$ plus ViT-B/16 and ViT-B/32 \cite{dosovitskiy2020image}. 
    \item \textbf{HtFE$^{txt}$$_2$}: fastText \cite{joulin2016bag} and Logistic Regression \cite{kleinbaum2002logistic}.
    \item \textbf{HtFE$^{txt}$$_4$}: HtFE$^{txt}$$_2$ plus LSTM \cite{hochreiter1997long} and BiLSTM \cite{schuster1997bidirectional}.
    \item \textbf{HtFE$^{txt}$$_{5\text{-}1}$}: Transformer models \cite{vaswani2017attention} with 1, 2, 4, 8, and 16 encoder layers, keeping 8 attention heads fixed. 
    \item \textbf{HtFE$^{txt}$$_{5\text{-}2}$}: Transformer models with 4 encoder layers and varying attention heads (1, 2, 4, 8, 16).
    \item \textbf{HtFE$^{txt}$$_{5\text{-}3}$}: Transformer models with encoder layers and heads scaling proportionally ((1,1), (2,2), (4,4), (8,8), (16,16)). 
    \item \textbf{HtFE$^{txt}$$_6$}: HtFE$^{txt}$$_4$ plus GRU \cite{cho2014learning} and Transformer (2 encoder layers, 8 heads) \cite{vaswani2017attention}. 
    \item \textbf{HtFE$^{sen}$$_2$}: HARCNNs \cite{zeng2014convolutional} with varying strides (1, 2).
    \item \textbf{HtFE$^{sen}$$_3$}: HARCNNs with varying strides (1, 2, 3).
    \item \textbf{HtFE$^{sen}$$_5$}: HtFE$^{sen}$$_3$ plus HARCNNs with 1 and 3 convolutional layers. 
    \item \textbf{HtFE$^{sen}$$_8$}: HARCNNs with 1, 2, and 3 convolutional layers and varying strides (1, 2, 3). 
\end{enumerate}
These models primarily differ in the feature extractor component, following existing HtFL works \cite{zhang2024fedtgp, zhang2024upload}. The feature extractor constitutes the main body of each model, typically employing various architectures, while the classifier part is usually a fully connected layer \cite{he2016deep}. More details are provided in the appendix and code.

\section{Benchmark Results of \ours}
\label{sec:results}

\begin{table*}[ht]
  \centering
  \caption{Test accuracy (\%) on Cifar100 under the Dirichlet setting with varying degrees of model heterogeneity. $\Delta$: The largest accuracy difference among HtFE$^{img}$$_2$, HtFE$^{img}$$_3$, HtFE$^{img}$$_4$, and HtFE$^{img}$$_9$.}
  \resizebox{!}{!}{
    \begin{tabular}{l|*{5}{c}|*{3}{c}}
    \toprule
    Settings & \multicolumn{5}{c|}{Heterogeneous Feature Extractors} & \multicolumn{3}{c}{Heterogeneous Models} \\
    \midrule
    & HtFE$^{img}$$_2$ & HtFE$^{img}$$_3$ & HtFE$^{img}$$_4$ & HtFE$^{img}$$_9$ & $\Delta$ & Res34-HtC$^{img}$$_4$ & HtFE$^{img}$$_8$-HtC$^{img}$$_4$ & HtM$^{img}$$_{10}$ \\
    \midrule
    LG-FedAvg & 46.61$\pm$0.24 & \ul{45.56$\pm$0.37} & \ul{43.91$\pm$0.16} & 42.04$\pm$0.26 & 4.57 & ---  & --- & --- \\
    FedGen & 43.92$\pm$0.11 & 43.65$\pm$0.43 & 40.47$\pm$1.09 & 40.28$\pm$0.54 & \ul{3.64} & ---  & --- & --- \\
    FedGH & \ul{46.70$\pm$0.35} & 45.24$\pm$0.23 & 43.29$\pm$0.17 & \ul{43.02$\pm$0.86} & 3.68 & ---  & --- & --- \\
    \midrule
    FML & 45.94$\pm$0.16 & 43.05$\pm$0.06 & 43.00$\pm$0.08 & 42.41$\pm$0.28 & 3.53 & 41.03$\pm$0.20 & 39.23$\pm$0.42 & 39.87$\pm$0.09 \\
    FedKD & 46.33$\pm$0.24 & 43.16$\pm$0.49 & 43.21$\pm$0.37 & 42.15$\pm$0.36 & 4.18 & 39.77$\pm$0.42 & 40.59$\pm$0.51 & 40.36$\pm$0.12 \\
    FedMRL & \ul{46.60$\pm$0.40} & \ul{44.50$\pm$0.60} & \ul{44.20$\pm$0.20} & \ul{43.90$\pm$0.40} & \textbf{\ul{2.70}} & \ul{45.79$\pm$0.42} & \ul{42.58$\pm$0.23} & \ul{42.10$\pm$0.10} \\
    \midrule
    FD & 46.88$\pm$0.13 & 43.53$\pm$0.21 & 43.56$\pm$0.14 & 42.09$\pm$0.20 & 4.79 & 44.72$\pm$0.13 & 41.67$\pm$0.06 & 40.95$\pm$0.04 \\
    FedProto & 43.97$\pm$0.19 & 38.14$\pm$0.64 & 34.67$\pm$0.55 & 32.74$\pm$0.82 & 11.23 & 32.26$\pm$0.19 & 25.57$\pm$0.72 & 36.06$\pm$0.10 \\
    FedTGP & \textbf{\ul{49.82$\pm$0.29}} & 49.65$\pm$0.37 & 46.54$\pm$0.14 & 48.05$\pm$0.19 & 3.28 & \textbf{\ul{48.19$\pm$0.27}} & \textbf{\ul{44.53$\pm$0.16}} & 41.91$\pm$0.21 \\
    FedKTL & 48.06$\pm$0.19 & \textbf{\ul{49.83$\pm$0.44}} & \textbf{\ul{47.06$\pm$0.21}} & \textbf{\ul{50.33$\pm$0.35}} & \ul{3.27} & 44.54$\pm$0.52 & 41.04$\pm$0.43 & \textbf{\ul{45.84$\pm$0.15}} \\
    \bottomrule
    \end{tabular}}
    \label{tab:hetero}
    \vspace{-5pt}
\end{table*}

We evaluate HtFL methods with image, text, and sensor signal tasks, analyzing their respective strengths and weaknesses, and highlight \ul{\textit{\textbf{key insights} in italics and underline}}. In each table, we use \textbf{bold} to highlight the best baseline among all counterparts, and \ul{underline} to indicate the best baseline within its respective category. 

\subsection{HtFL with Image}

\subsubsection{\textbf{Performance in Label Skew Settings}}

In \cref{tab:datasets_acc}, we first evaluate three categories of HtFL methods and analyze their performance on four popular benchmark datasets. 
The results indicate that 
(1) FedTGP outperforms all baselines in most cases, demonstrating its practical adaptability. This highlights that \ul{\textit{discriminability-improved lightweight prototypes are an effective solution for HtFL on image tasks}}. (2) Among partial parameter sharing methods, FedGH outperforms other methods, highlighting \ul{\textit{the effectiveness of calibrating the global classifier using local prototypes}}. (3) In mutual distillation methods, FedMRL performs better than other baselines, as it leverages both the auxiliary and local heterogeneous models to extract features during inference, thereby enriching the local model's capabilities. (4) Among prototype-sharing methods, FedKTL also shows superiority in many cases, illustrating that \ul{\textit{using image-prototype pairs to augment the original prototypes can bring additional benefits for knowledge transfer among heterogeneous clients on image tasks}}. (5) Mutual distillation generally outperforms partial parameter sharing across methods and datasets in the Dirichlet setting, while prototype sharing exhibits variability among methods. 

\subsubsection{\textbf{Performance in the Feature Shift Setting}}

\begin{figure}
    \centering    \includegraphics[width=\linewidth]{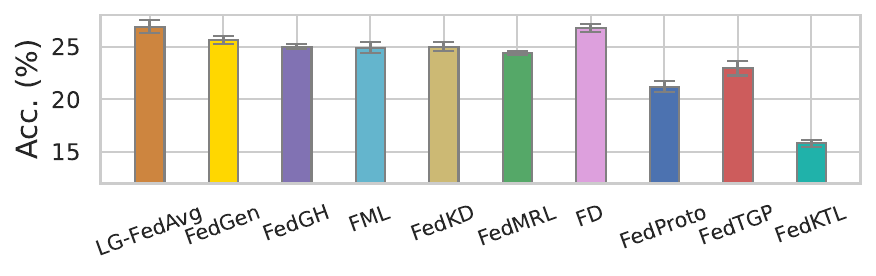}
    \vspace{-0.3cm}
    \vspace{-5pt}
    \caption{Test accuracy (\%) on DomainNet under the feature shift scenario using HtFE$^{img}$$_4$.}
    \vspace{-0.3cm}
    \label{fig: DomainNet}
    \vspace{-5pt}
\end{figure}

From \cref{fig: DomainNet}, we observe that LG-FedAvg and FD show superior results. \ul{\textit{The diverse features in this scenario exacerbate the challenge of aligning the feature space for prototype-sharing methods}} like FedProto, FedTGP, and FedKTL. 
Among these, FedKTL shows a significant performance gap, as the pre-trained generator 
primarily generates real-world images, which do not align well with the clipart, sketch, infographic, painting, and quickdraw images in DomainNet. 

\subsubsection{\textbf{Impact of Model Heterogeneity}}

With diverse \textit{heterogeneous feature extractors} in \cref{tab:hetero}, we observe that (1) most HtFL methods show decreased accuracy as model heterogeneity increases, while \ul{\textit{FedMRL is the most robust among them. Specifically, FedMRL benefits from its combination of auxiliary global and local models}}, resulting in an accuracy difference ($\Delta$) of 2.70\% from HtFE$^{img}$$_2$ to HtFE$^{img}$$_9$, compared to a 3.27\%–11.23\% difference for other baselines. 
(2) Among prototype sharing methods, FD and FedProto share prototypes in the logit and feature space, respectively. FedProto's performance lags behind FD, especially in highly heterogeneous settings. The accuracy gap is 2.91\% for HtFE$^{img}$$_2$, but widens to 9.35\% with HtFE$^{img}$$_9$, as feature extraction is more affected than logit prediction with heterogeneous feature extractors. 

We then further introduce \textit{heterogeneous models} where the classifier part is also heterogeneous, making partial parameter sharing methods inapplicable here. As shown in \cref{tab:hetero}, (1) FedTGP maintains its superiority across diverse heterogeneous model settings due to its \ul{\textit{adaptive refinement of prototypes, making it less sensitive to heterogeneous classifiers.}} (2) Among mutual distillation methods, FedKD performs the worst with Res34-HtC$^{img}$$_4$, but ranks second in HtFE$^{img}$$_8$-HtC$^{img}$$_4$ and HtM$^{img}$$_{10}$, highlighting the \ul{\textit{advantage of aligning feature vectors with heterogeneous feature extractors over homogeneous ones}}. 

\begin{table}[ht]
    \centering
    \caption{Test accuracy (\%) on Cifar100 in the Dirichlet setting using HtFE$^{img}$$_8$ with different values of $\alpha$. The results in ``()'' indicate the total number of converged rounds. We omit error bars here due to limited space. }
    \resizebox{\linewidth}{!}{
    \begin{tabular}{l|ccccccc}
    \toprule
     & \(\alpha = 0.01\) & \(\alpha = 0.1\) & \(\alpha = 0.5\) & \(\alpha = 1\) \\
    \midrule
    LG-FedAvg & \ul{66.62 (178)} & 40.65 (190) & \ul{21.32 (273)} & \ul{15.73 (141)} \\
    FedGen & 66.61 (153) & 38.73 (152) & 21.19 (144) & 15.41 (153) \\
    FedGH & 65.23 (146) & \ul{40.99 (226)} & 21.21 (232) & 15.53 (194) \\
    \midrule
    FML & 64.53 (370) & 39.86 (287) & 20.05 (150) & 16.02 (319) \\
    FedKD & 64.93 (285) & 40.56 (198) & 21.52 (166) & \ul{16.34 (288)} \\
    FedMRL & \ul{68.82 (191)} & \ul{41.20 (170)} & \ul{22.33 (152)} & 16.32 (567)  \\
    \midrule
    FD & 67.01 (338) & 41.54 (216) & 22.13 (161) & 16.42 (273) \\
    FedProto & 60.62 (540) & 36.34 (533) & 19.34 (570) & 12.63 (369) \\
    FedTGP  & 69.28 (237) & 46.94 (211) & 21.80 (220) & 19.03 (279) \\
    FedKTL & \textbf{\ul{71.25 (138)}} & \textbf{\ul{46.94 (152)}} & \textbf{\ul{25.06 (141)}} & \textbf{\ul{19.91 (122)}} \\
    \bottomrule
    \end{tabular}}
    \label{tab:alpha}
    
\end{table}

\subsubsection{\textbf{Impact of Data Heterogeneity}}

HtFL considers both data and model heterogeneity. To further investigate HtFL methods under varying data heterogeneity together with model heterogeneity, we conducted additional experiments using HtFE$^{img}$$_8$ and $\alpha$ values of 0.01, 0.5, and 1, as shown in \cref{tab:alpha}. FedKTL outperforms other baselines in all settings, as its data augmentation approach can alleviate the effect of data heterogeneity. 

Regarding convergence rate, we find that the \ul{\textit{convergence behavior of most baselines is significantly affected by data heterogeneity}}, with \ul{\textit{FedGen, FedTGP, and FedKTL demonstrating stable convergence rates}}. In terms of total convergence rounds, methods like FedMRL and FedProto require considerably more rounds at certain $\alpha$ values. Specifically, FedMRL requires 567 rounds for $\alpha=1$, while FedProto requires 540, 533, and 570 rounds for $\alpha=0.01$, $\alpha=0.1$, and $\alpha=1$, respectively. Among all methods, FedGen, FedKD, and FedKTL converge the fastest. 

\begin{table}[ht]
  \centering
  \caption{Test accuracy (\%) on Cifar100 in the Dirichlet setting using HtFE$^{img}$$_8$ with a large number of clients.}
  \resizebox{\linewidth}{!}{
    \begin{tabular}{l|ccc|c}
    \toprule
    & \multicolumn{3}{c|}{$\rho=50\%$} & $\rho=10\%$ \\
    \midrule
    & 50 Clients & 100 Clients & 200 Clients & 100 Clients \\
    \midrule
    LG-FedAvg & 37.81$\pm$0.12 & \ul{35.14$\pm$0.47} & 27.93$\pm$0.04 & \ul{41.01$\pm$0.29} \\
    FedGen & \ul{37.95$\pm$0.25} & 34.52$\pm$0.31 & 28.01$\pm$0.24 & 34.30$\pm$0.51\\
    FedGH & 37.30$\pm$0.44 & 34.32$\pm$0.16 & \ul{29.27$\pm$0.39} & 40.34$\pm$0.81 \\
    \midrule
    FML & 38.47$\pm$0.14 & 36.09$\pm$0.28 & 30.55$\pm$0.52 & 35.24$\pm$0.91 \\
    FedKD & 38.25$\pm$0.41 & 35.62$\pm$0.55 & \ul{31.82$\pm$0.50} & 36.53$\pm$0.27 \\
    FedMRL & \ul{38.60$\pm$0.20} & \ul{36.40$\pm$0.60} & 30.66$\pm$0.78 & \textbf{\ul{41.70$\pm$0.30}} \\
    \midrule
    FD & 38.51$\pm$0.36 & 36.06$\pm$0.24 & 31.26$\pm$0.13 & \ul{41.23$\pm$0.53} \\
    FedProto & 33.03$\pm$0.42 & 28.95$\pm$0.51 & 24.28$\pm$0.46 & 28.64$\pm$0.95 \\
    FedTGP & \textbf{\ul{43.17$\pm$0.23}} & \textbf{\ul{41.57$\pm$0.30}} & 32.28$\pm$0.68 & 32.53$\pm$0.51\\
    FedKTL & 43.16$\pm$0.82 & 39.73$\pm$0.87 & \textbf{\ul{34.24$\pm$0.45}} & 37.61$\pm$0.42 \\
    \bottomrule
    \end{tabular}}
    \label{tab:largeclient}
\end{table}

\subsubsection{\textbf{Impact of Client Participation Ratio with More Clients}}
We evaluate the baselines across three scenarios with 50, 100, and 200 clients to assess the scalability of each baseline with a large number of clients and partial participation ratio per round ($\rho<100\%$). From \cref{tab:largeclient}, we observe that:
(1) All baselines exhibit reduced performance as the number of clients increases. This is due to the smaller amount of data available per client when Cifar100 is distributed across more clients, leading to a decline in performance for all methods. 
(2) The combination of the small global model and the local model in FedMRL helps mitigate the insufficient knowledge for aggregation caused by partial participation, particularly at low $\rho$. 
(3) While FedTGP and FedKTL perform well with 100 clients and $\rho=50\%$, their performance drops with $\rho=10\%$, especially for FedTGP. \ul{\textit{With lower client participation, FedTGP struggles to aggregate enough knowledge from clients in each round}}, leading to poor prototype guidance during local training. (4) In contrast, \ul{\textit{FedKTL, with its pre-trained large generator, can replenish knowledge to the prototypes, mitigating the issue of insufficient knowledge}}. Thanks to this knowledge replenishment feature, FedKTL also performs well in scenarios with more clients, such as with 200 clients. \ul{\textit{This suggests promising future work on integrating HtFL frameworks with pre-trained large models (PLMs) for large-scale scenarios.}}

\begin{figure}
    \centering    \includegraphics[width=\linewidth]{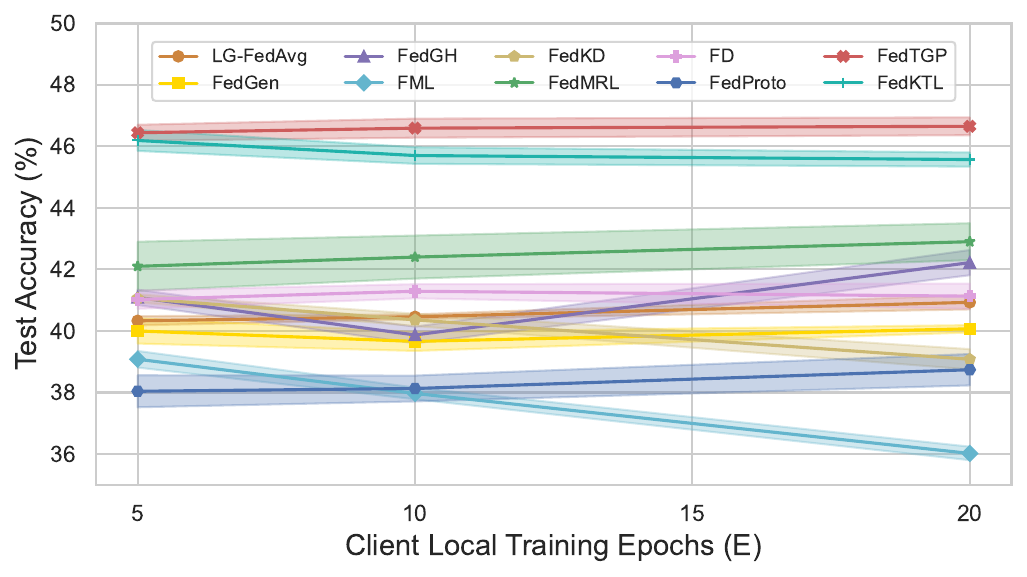}
    \vspace{-0.3cm}
    \caption{Test accuracy (\%) on Cifar100 in the Dirichlet setting using HtFE$^{img}$$_8$ with a large local training epochs $E$.}
    \vspace{-0.3cm}
    \label{fig: Epochs}
\end{figure}

\subsubsection{\textbf{Impact of Local Training Epochs}}

Multiple local training epochs ($E$) on the client during FL training can help reduce the communication burden \cite{mcmahan2017communication}. In \cref{fig: Epochs}, prototype-sharing methods maintain their maximum performance. However, \ul{\textit{in the case of mutual distillation methods like FML and FedKD, increasing the number of training epochs leads to a decrease in performance}}. This is because both methods rely on an auxiliary model, and as $E$ increases, the auxiliary model accumulates more biased information during training, which can negatively impact model aggregation. In contrast, FedMRL alleviates this issue by merging the features extracted by the auxiliary and local models. 

\begin{figure}[h]
    \centering \includegraphics[width=\linewidth]{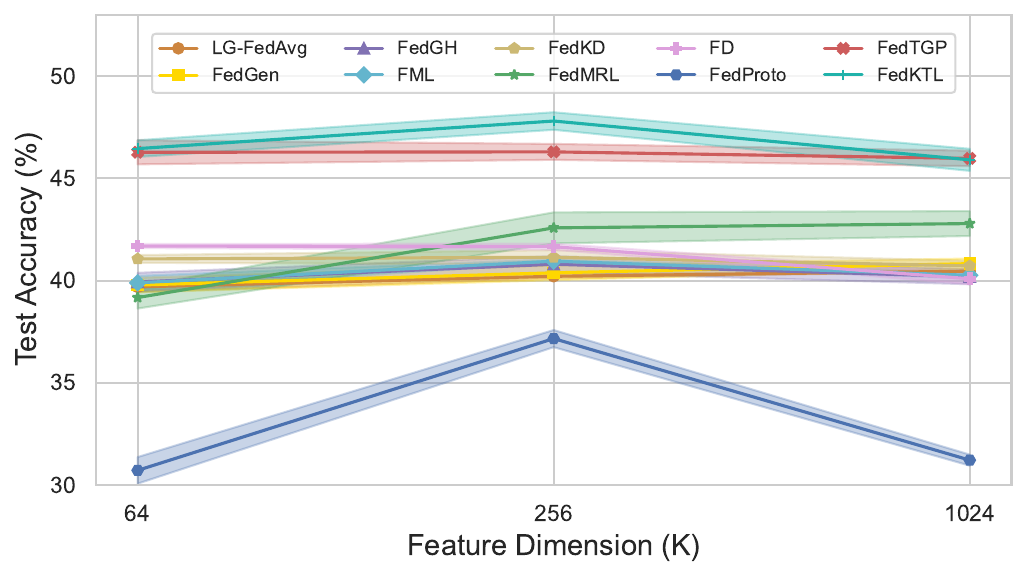}
    \vspace{-0.3cm}
    \caption{Test accuracy (\%) on Cifar100 in the Dirichlet setting using HtFE$^{img}$$_8$ with varying feature dimensions $K$.}
    \vspace{-0.3cm}
    \label{fig: DimK}
\end{figure}

\subsection{Impact of Feature Dimensions}

In \cref{fig: DimK}, we observe that most methods show improved performance as the number of feature dimension $K$ increases from 64 to 256. However, methods that share partial model parameters, such as LG-FedAvg and FedGen, do not follow this trend. All other methods achieve their best performance at $K=256$. 
All methods show consistent or improved performance as \( K \) increases from 64 to 256, but for some methods like FD, FedProto, and FedKTL, a very high \( K \) may lead to a performance drop.

\begin{table}[ht]
  \centering
  \caption{The communication and computation costs on Cifar100 in the default Dirichlet setting using HtFE$^{img}$$_8$. ``MB'' and ``s'' stand for megabytes and seconds, respectively. }
  \resizebox{!}{!}{
    \begin{tabular}{l|*{2}{c}|*{2}{c}}
    \toprule
    Items & \multicolumn{2}{c|}{Comm. (MB)} & \multicolumn{2}{c}{Computation (s)} \\
    \midrule
     & Up. & Down. & Client & Server \\
    \midrule
    LG-FedAvg & 3.93 & 3.93 & 6.19 & 0.04 \\
    FedGen & 3.93 & 29.22 & 5.77 & 2.96 \\
    FedGH & 1.75 & 3.93 & 9.53 & 0.37 \\
    \midrule
    FML & 70.57 & 70.57 & 8.63 & 0.07 \\
    FedKD & 63.02 & 63.02 & 9.04 & 0.07 \\
    FedMRL & 70.57 & 70.57 & 9.14 & 0.07 \\
    \midrule
    FD & 0.34 & 0.76 & 6.52 & 0.03 \\
    FedProto & 1.75 & 3.89 & 6.65 & 0.04 \\
    FedTGP & 1.75 & 3.89 & 6.55 & 7.87 \\
    FedKTL & 0.34 & 27.35 & 8.92 & 8.95 \\
    \bottomrule
    \end{tabular}}
    \label{tab:three}
    \vspace{-5pt}
\end{table}

\subsubsection{\textbf{Communication Costs}}
We calculate the communication overhead as the total upload and download bytes from all participating clients in each round, using the float32 data type (4 bytes per number) in PyTorch \cite{paszke2019pytorch}. From \cref{tab:three}, we observe that: (1) \ul{\textit{Although the mutual distillation method transmits a relatively small global model, its communication costs remain high}}. The use of \ul{\textit{singular value decomposition (SVD) does not significantly reduce the communication overhead}} in FedKD. (2) \ul{\textit{Most prototype-sharing methods require minimal upload/download bytes due to the lightweight nature of the prototypes}}, while FedKTL incurs additional communication costs by augmenting prototypes with corresponding images. 

\subsubsection{\textbf{Computation Costs}}
To evaluate the execution of basic operations, we calculate the average GPU execution time for each client and server on idle GPUs in each round, presenting this as the time cost in \cref{tab:three}. The results show that: (1) \ul{\textit{Mutual distillation methods incur higher client training time due to the additional auxiliary model learning}}. (2) Methods that only use the server for averaging require minimal server costs. (3) FedGen, FedTGP, and FedKTL involve extra server-side training and multiple rounds, leading to higher computational power consumption on the server compared to other baselines. 

\begin{table}[ht]
    \centering
    \caption{Test accuracy (\%) on three medical datasets: KVASIR (HtFE$^{img}$$_8$), COVIDx (HtM$^{img}$$_{10}$) and Camelyon17 (HtFE$^{img}$$_5$). }
    \resizebox{\linewidth}{!}{
    \begin{tabular}{l|cc|c}
    \toprule
     & \multicolumn{2}{c|}{Dirichlet Setting} & Real-World Setting\\
    \midrule
    Data & KVASIR \cite{pogorelov2017kvasir} & COVIDx \cite{wang2020covid} & Camelyon17 \cite{koh2021wilds} \\
    \midrule
    \textcolor{gray_}{Pre-trained} & \textcolor{gray_}{26.52} & \textcolor{gray_}{37.60} & \textcolor{gray_}{66.81} \\
    FML & \textbf{27.24} (\textcolor{green_}{\textbf{+0.72}}) & 39.57 (\textcolor{green_}{\textbf{+1.97}}) & 68.76 (\textcolor{green_}{\textbf{+1.95}}) \\
    FD & 26.78 (\textcolor{green_}{\textbf{+0.26}}) & \textbf{40.02} (\textcolor{green_}{\textbf{+2.42}}) & \textbf{69.12} (\textcolor{green_}{\textbf{+2.31}}) \\
    \bottomrule
    \end{tabular}}
    \label{tab:medical}
    \vspace{-10pt}
\end{table}

\begin{figure}[h]
    \centering
    \includegraphics[width=\linewidth]{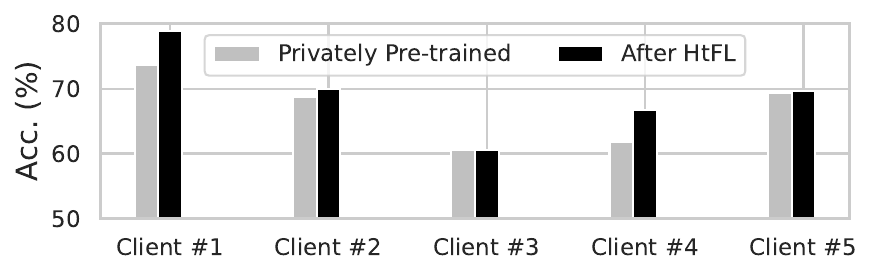}
    \vspace{-0.3cm}
    \caption{Test accuracy (\%) per client on the real-world Camelyon17 dataset using FD, where 5 hospitals each own a distinct heterogeneous model from HtFE$^{img}$$_5$.}
    \vspace{-0.3cm}
    \label{fig:clients}
\end{figure}

\subsubsection{\textbf{Performance on Medical Datasets with Black-boxed Pre-trained Heterogeneous Models}}
Here, we present a \textbf{realistic application} that illustrates the value of heterogeneous model collaborative learning: hospitals that have developed and \textit{locally pre-trained their models for specific needs but face limitations due to insufficient local data}. By collaborating with other hospitals in the same field, they can \textbf{\textit{further improve}} their heterogeneous models. This scenario is especially common among medium and small institutions, as AI adoption has been ongoing for years, and \textit{many organizations already have their unique models in place} \cite{castiglioni2021ai, cao2022ai, khalil2021deep}. 

\textit{We first privately pre-train the heterogeneous models locally until convergence and then apply HtFL methods for post-training.} Here, we focus on the generalization ability of client models, a key interest in the medical field \cite{gao2024desam, guan2021domain}, and evaluate them on a global test set, as shown in \cref{tab:medical}, where we assign 5 heterogeneous models from HtFE$^{img}$$_5$ to the 5 hospitals in Camelyon17, respectively. 

\textbf{Experimental Results.} The results show that \ul{\textit{HtFL further enhances the quality of heterogeneous black-box models compared with pre-trained models, demonstrating broader utility.}} Besides, sharing prototypes like FD mostly gains more improvements than sharing an auxiliary tiny model in FML. The results in \cref{fig:clients} further demonstrate that \ul{\textit{HtFL can enhance the quality of the pre-trained black-box model for each participating client}}. This \ul{\textit{realistic black-box model setting is under-explored in the literature, with only a few methods applicable, highlighting the need for future research.}} 

\subsection{HtFL with Text}
In this section, we compare various methods in the text modality. Note that FedKTL is excluded as it is limited to image tasks. 

\subsubsection{\textbf{Performance on Various Data Heterogeneity Scenarios.}}\label{sec:text with data hetero}

\begin{table}[ht]
  \centering
  \caption{Test accuracy (\%) on AG News and Shakespeare using HtFE$^{txt}$$_6$.}
  \resizebox{!}{!}{
    \begin{tabular}{l|cc|c}
    \toprule
     & \multicolumn{2}{c|}{AG News} & Shakespeare \\
    \midrule
    Scenarios & Pathological & Dirichlet & Real-World \\
    \midrule
    LG-FedAvg & 52.52$\pm$0.04 & 71.89$\pm$0.20 & 55.87$\pm$0.52 \\
    FedGen & 57.08$\pm$0.11 & 77.16$\pm$0.25 & \textbf{\ul{57.18$\pm$0.31}} \\
    FedGH & \textbf{\ul{64.01$\pm$0.28}} & \ul{79.72$\pm$0.19} & 49.81$\pm$0.47 \\
    \midrule
    FML & 54.33$\pm$0.13 & 83.13$\pm$0.21 & 49.62$\pm$0.24 \\
    FedKD & 56.39$\pm$0.27 & 88.62$\pm$0.05 & \ul{50.08$\pm$0.62} \\
    FedMRL & \ul{57.01$\pm$0.05} & \textbf{\ul{88.69$\pm$0.16}} & 42.49$\pm$0.54 \\
    \midrule
    FD & \ul{60.35$\pm$0.02} & \ul{87.73$\pm$0.17} & \ul{35.46$\pm$0.13} \\
    FedProto & 38.55$\pm$0.12 & 47.16$\pm$0.15 & 13.15$\pm$0.17 \\
    FedTGP & 45.42$\pm$0.23 & 64.70$\pm$0.19 & 32.67$\pm$0.44 \\
    \bottomrule
    \end{tabular}}
    \label{tab:text with data hetero}
    
\end{table}

We consider three heterogeneous scenarios in the text modality and conduct 100 rounds for all baselines, utilizing the HtFE$^{txt}$$_6$ model group, which has the highest degree of model heterogeneity. From \cref{tab:text with data hetero}, we observe the following key findings: 
(1) Although FedMRL achieves the best performance among mutual distillation methods in label skew scenarios, its advantages vanish in the real-world scenario.
(2) Given text data, FedProto and FedTGP perform relatively poorly compared to image tasks. \ul{\textit{This suggests that in the text domain, models with different architectures have significant differences in their processing mechanisms, feature extraction strategies, and context modeling capabilities, making it difficult to align their outputs into a unified representation space.}} In contrast, \ul{\textit{aligning clients in the logit space proves to be more efficient and effective than feature-space alignment}}. 
Addressing this challenge at the prototype level remains an open research problem. 

\begin{table}[ht]
  \centering
  \caption{Test accuracy (\%) on AG News in the Dirichlet settings with various model heterogeneity.}
  \resizebox{!}{!}{
    \begin{tabular}{l|*{3}{c}}
    \toprule
     & HtFE$^{txt}$$_{2}$ & HtFE$^{txt}$$_{4}$ & HtFE$^{txt}$$_{6}$ \\
    \midrule
    LG-FedAvg & 83.63$\pm$0.09 & 74.69$\pm$0.24 & 71.89$\pm$0.20 \\
    FedGen & 83.53$\pm$0.07 & \ul{81.30$\pm$0.29} & 77.16$\pm$0.25 \\
    FedGH & \ul{85.35$\pm$0.02} & 77.04$\pm$0.24 & \ul{79.72$\pm$0.19} \\
    \midrule
    FML & 81.83$\pm$0.07 & 85.92$\pm$0.14 & 83.13$\pm$0.21 \\
    FedKD & \ul{88.14$\pm$0.01} & \textbf{\ul{88.06$\pm$0.27}} & 88.62$\pm$0.05 \\
    FedMRL & 85.72$\pm$0.12 & 87.69$\pm$0.19 & \textbf{\ul{88.69$\pm$0.16}} \\
    \midrule
    FD & \textbf{\ul{91.35$\pm$0.14}} & \ul{79.06$\pm$0.25} & 8\ul{7.73$\pm$0.17} \\
    FedProto & 52.88$\pm$0.04 & 35.66$\pm$0.19 & 47.16$\pm$0.15 \\
    FedTGP & 47.11$\pm$0.14 & 62.97$\pm$0.21 & 64.70$\pm$0.19 \\
    \bottomrule
    \end{tabular}}
    \label{tab:text with model hetero}
\end{table}

\subsubsection{\textbf{Impact of Model Heterogeneity.}}\label{sec:text with model hetero}
According to the results in \cref{tab:text with model hetero}, we observe the following: 
(1) For partial parameter-sharing methods, performance generally degrades as model heterogeneity increases. 
(2) In contrast, mutual distillation and prototype-sharing methods do not exhibit a strictly negative correlation with heterogeneity. (3) Besides the heterogeneity among models, the quality of feature extraction plays a crucial role in prototype-based methods. In HtFE$^{txt}$$_{4}$ and HtFE$^{txt}$$_{6}$, stronger feature extraction models are gradually introduced, improving the quality of prototypes. 

\begin{table}[ht]
  \centering
  \caption{Test accuracy (\%) on AG News in the Dirichlet settings with Transformer models.}
  \resizebox{!}{!}{
    \begin{tabular}{l|*{3}{c}}
    \toprule
     & HtFE$^{txt}$$_{5\text{-}1}$ & HtFE$^{txt}$$_{5\text{-}2}$ & HtFE$^{txt}$$_{5\text{-}3}$ \\
    \midrule
    LG-FedAvg & \ul{96.18$\pm$0.06} & \ul{96.17$\pm$0.06} & 95.86$\pm$0.07 \\
    FedGen & 95.99$\pm$0.14 & 95.96$\pm$0.06 & 95.70$\pm$0.05 \\
    FedGH & 95.76$\pm$0.02 & 95.88$\pm$0.13 & \ul{95.88$\pm$0.06} \\
    \midrule
    FML & \textbf{\ul{96.57$\pm$0.01}} & \textbf{\ul{96.52$\pm$0.05}} & \textbf{\ul{96.31$\pm$0.04}} \\
    FedKD & 96.10$\pm$0.07 & 95.20$\pm$0.01 & 95.40$\pm$0.10 \\
    FedMRL & 96.06$\pm$0.14 & 95.95$\pm$0.09 & 95.85$\pm$0.07 \\
    \midrule
    FD & \ul{96.10$\pm$0.13} & \ul{96.17$\pm$0.11} & 95.99$\pm$0.13 \\
    FedProto & 95.91$\pm$0.08 & 95.92$\pm$0.04 & 95.85$\pm$0.04 \\
    FedTGP & 96.04$\pm$0.08 & 95.93$\pm$0.06 & \ul{96.04$\pm$0.12} \\
    \bottomrule
    \end{tabular}}
    \label{tab:text Transformers}
\end{table}

\subsubsection{\textbf{Performance on Transformer Models}}\label{sec:text with Transformer}

Recently, Transformer models have demonstrated exceptional capabilities across various tasks, particularly in the text modality \cite{touvron2023llama, achiam2023gpt, liu2024deepseek}. In \cref{tab:text Transformers}, we explore collaborative learning among heterogeneous Transformer models. With powerful Transformer architectures, the performance of all baselines improves significantly compared to \cref{tab:text with model hetero}. Moreover, they show increased robustness to varying model heterogeneity, with minimal performance differences. This suggests that \ul{\textit{strong model capabilities in client models enable effective collaboration across different HtFL methods despite model heterogeneity}}. 

\subsection{HtFL with Sensor Signal}

\begin{table}[ht]
  \centering
  \caption{Test accuracy (\%) on HAR and PAMAP2 in the real-world setting using HtFE$^{sen}$$_8$.}
  \resizebox{!}{!}{
    \begin{tabular}{l|*{2}{c}}
    \toprule
     & HAR & PAMAP2 \\
    \midrule
    LG-FedAvg & \ul{94.64$\pm$0.14} & \ul{92.71$\pm$0.11} \\
    FedGen & 93.98$\pm$0.25 & 91.36$\pm$0.04 \\
    FedGH & 94.25$\pm$0.14 & 90.11$\pm$0.06 \\
    \midrule
    FML & 94.58$\pm$0.13 & 90.78$\pm$0.10 \\
    FedKD & \ul{95.27$\pm$0.15} & \textbf{\ul{94.40$\pm$0.02}} \\
    FedMRL & 94.34$\pm$0.24 & 91.44$\pm$0.33 \\
    \midrule
    FD & \textbf{\ul{95.71$\pm$0.01}} & \ul{91.34$\pm$0.02} \\
    FedProto & 92.01$\pm$0.63 & 84.17$\pm$0.02 \\
    FedTGP & 90.11$\pm$1.69 & 76.99$\pm$0.11 \\
    \bottomrule
    \end{tabular}}
    \label{tab:sen acc}
\end{table}

\subsubsection{\textbf{Performance on Different Datasets}}
We study real-world sensor signal modality using the highly heterogeneous HtFE$^{sen}$$_8$ model group with 500 rounds for all methods. 
Since PAMAP2 covers a broader range of physical activities and continuous sensor data from multiple body parts compared to HAR, it is more complex. As shown in \cref{tab:sen acc}, (1) \ul{\textit{HtFL methods perform well on simpler sensor signal tasks, but their performance declines as task complexity increases}}.
(2) Prototype-sharing methods experience a more significant decline. Specifically, FedProto and FedTGP show drops of 7.84\% and 13.12\%, respectively, while other methods experience a decline of only 0.87\% to 3.80\%. This is due to the nature of prototypes, which are averages of class representations. \ul{\textit{While effective for the static image modality, prototypes struggle to capture the continuous and dynamic nature of sensor signal, where temporal dependencies and noise hinder meaningful representation.}} 
(3) Mutual distillation methods, such as FedKD, perform best across all categories, demonstrating that \ul{\textit{sharing a well-structured, homogeneous auxiliary model is better suited for handling continuous and dynamic data}}, enabling more effective knowledge transfer across clients. 

\begin{table}[ht]
  \centering
  \caption{Test accuracy (\%) on HAR in the real-world setting using different model groups.}
  \resizebox{!}{!}{
    \begin{tabular}{l|*{3}{c}}
    \toprule
     & HtFE$^{sen}$$_2$ & HtFE$^{sen}$$_3$ & HtFE$^{sen}$$_5$ \\
    \midrule
    LG-FedAvg & 94.62$\pm$0.01 & 94.60$\pm$0.02 & \ul{94.72$\pm$0.06} \\
    FedGen & \ul{94.86$\pm$0.17} & \ul{94.99$\pm$0.04} & 93.73$\pm$0.11 \\
    FedGH & 94.23$\pm$0.05 & 94.28$\pm$0.01 & 94.06$\pm$0.12 \\
    \midrule
    FML & 94.86$\pm$0.20 & 94.95$\pm$0.11 & 94.58$\pm$0.08 \\
    FedKD & \ul{95.70$\pm$0.54} & \ul{96.07$\pm$0.03} & \ul{95.39$\pm$0.08} \\
    FedMRL & 94.59$\pm$0.39 & 94.77$\pm$0.19 & 94.32$\pm$0.22 \\
    \midrule
    FD & 95.76$\pm$0.02 & 95.75$\pm$0.03 & \textbf{\ul{95.70$\pm$0.01}} \\
    FedProto & 95.44$\pm$0.47 & 95.79$\pm$0.03 & 92.44$\pm$0.03 \\
    FedTGP & \textbf{\ul{96.73$\pm$0.42}} & \textbf{\ul{97.03$\pm$0.12}} & 91.31$\pm$0.11 \\
    \bottomrule
    \end{tabular}}
    \label{tab:sen hetero}
\end{table}

\subsubsection{\textbf{Impact of Model Heterogeneity}}
We vary the degree of model heterogeneity by adjusting the strides and the number of convolutional layers in the HARCNN \cite{zeng2014convolutional} to assess their impact on HtFL methods' performance. 
(1) As shown in \cref{tab:sen hetero}, HtFL methods perform better with varying strides in models with homogeneous convolutional layers but worsen in models with heterogeneous layers. This is because varying strides improve the model's ability to extract features at different scales, while changes in convolutional layers increase feature dimensionality, leading to higher complexity and reduced generalization. (2) Among the prototype-sharing methods, FedTGP performs best in HtFE$^{sen}$$_2$ and HtFE$^{sen}$$_3$, but worst in HtFE$^{sen}$$_5$, due to the impact of high feature dimensionality.

\section{Conclusion and Future Directions}

In this work, we introduce \ours, an easy-to-use, versatile, and extensible framework that provides a comprehensive benchmark for both research and practical applications in HtFL. 
\ours's support for heterogeneous models in collaborative learning opens promising future directions, particularly in incorporating complex \textit{pre-trained large models}, \textit{black-box models}, and other \textit{diverse models from different tasks and modalities}.

\section{Acknowledgments}

This work was supported by the National Key R\&D Program of China (Grant No.2022ZD0160504) and the Interdisciplinary Program of Shanghai Jiao Tong University (project No.YG2024QNB05).

\bibliographystyle{ACM-Reference-Format}
\balance
\bibliography{main}

\appendix

\section{Additional Experimental Details}

\subsection{Training and evaluation details}

By default, we divide each client's dataset into a training set and a test set with a 3:1 split and report the average test accuracy across all clients' test sets. 
In line with standard practices \cite{mcmahan2017communication}, we perform one local training epoch per communication round, using a batch size of 10, which corresponds to $\lfloor \frac{k_i}{10} \rfloor$ update SGD \cite{zhang2015deep} steps. Each experiment (default: 1000 rounds) is repeated three times with a client learning rate of 0.01. We report the best results along with error bars. By default, we consider full client participation ($\rho=100\%$) using 20 clients, while adopting partial participation ($\rho \le 50\%$) for scenarios with large client counts, such as 200 clients. 

\subsection{Experimental environment}

We experimented on a machine equipped with 64 Intel(R) Xeon(R) Platinum 8362 CPUs, 256 GB of memory, 8 NVIDIA 3090 GPUs, and running Ubuntu 20.04.4 LTS. Typically, our experiments are completed within 48 hours. However, those involving a large number of clients and extensive local training epochs may require up to a week to finish. 

\subsection{Heterogeneous Model Architectures}

As we use existing model architectures for image tasks, we only list the specific models for text and sensor signals here. 

\subsubsection{Text Modality Model} 
\begin{itemize}[left=0pt, labelsep=0.6em]
    \item \textbf{Architectures in HtFE$^{txt}$$_2$}: This model group combines fastText \cite{joulin2016bag} and Logistic Regression \cite{kleinbaum2002logistic}.
    \begin{enumerate}[left=0pt, labelsep=0.6em]
        \item fastText: This model uses an embedding layer followed by a linear hidden layer and a final output layer. 
        \item Logistic Regression: This is a traditional linear classifier applied directly to the word embeddings.
    \end{enumerate}
    \item \textbf{Architectures in HtFE$^{txt}$$_4$}: This model group extends HtFE$^{txt}$$_2$ by adding LSTM \cite{hochreiter1997long} and BiLSTM \cite{schuster1997bidirectional} models.
    \begin{enumerate}[left=0pt, labelsep=0.6em]
        \item LSTM: This model uses an embedding layer, followed by 2 LSTM layers, and a fully connected output layer.
        \item BiLSTM: Similar to LSTM, but with a bidirectional LSTM layer.
    \end{enumerate}
    \item \textbf{Architectures in HtFE$^{txt}$$_{5\text{-}1}$}: This model group uses Transformer \cite{vaswani2017attention} models with varying numbers of encoder layers, specifically 1, 2, 4, 8, and 16 layers.
    \begin{enumerate}[left=0pt, labelsep=0.6em]
        \item All Transformer models keep the number of attention heads fixed at 8.
        \item Each model consists of an embedding layer, multiple transformer encoder layers, and a final fully connected classification layer.
    \end{enumerate}
    \item \textbf{Architectures in HtFE$^{txt}$$_{5\text{-}2}$}: This model group is similar to HtFE$^{txt}$$_{5\text{-}1}$, but here the number of attention heads is varied (1, 2, 4, 8, 16), with the number of encoder layers fixed at 4.
    \item \textbf{Architectures in HtFE$^{txt}$$_{5\text{-}3}$}: This model group is similar to HtFE$^{txt}$$_{5\text{-}1}$. The encoder layers and attention heads scale in pairs, such as (1,1), (2,2), (4,4), (8,8), and (16,16). 
    \item \textbf{Architectures in HtFE$^{txt}$$_6$}: This model group extends HtFE$^{txt}$$_4$ by adding GRU \cite{cho2014learning} and Transformer \cite{vaswani2017attention} models.
    \begin{enumerate}[left=0pt, labelsep=0.6em]
        \item GRU: This model uses an embedding layer, followed by 2 GRU layers, and a fully connected output layer.
        \item Transformer: This model consists of an embedding layer, 2 transformer encoder layers with 8 attention heads, and a final classification layer.
    \end{enumerate}
\end{itemize}

\subsubsection{Sensor signal Modality Model} 

\begin{itemize}[left=0pt, labelsep=0.6em]
    \item \textbf{Architectures in HtFE$^{sen}$$_2$}: This model group uses HARCNNs \cite{zeng2014convolutional} with varying strides (1, 2).
    \begin{enumerate}[left=0pt, labelsep=0.6em]
        \item HARCNN: The model consists of 2 convolutional layers, followed by 2 pooling layers and 3 fully connected layers.
        \item The strides of the convolutional layers are set to 1 and 2.
    \end{enumerate}
    \item \textbf{Architectures in HtFE$^{sen}$$_3$}: This model group is similar to HtFE$^{sen}$$_2$, but the strides of the convolutional layers are varied to 1, 2, and 3.
    \item \textbf{Architectures in HtFE$^{sen}$$_5$}: This model group builds on HtFE$^{sen}_3$ by varying the number of convolutional layers.
    \begin{enumerate}[left=0pt, labelsep=0.6em]
        \item HARCNN1 with 1 convolutional layer: The model consists of 1 convolutional layer, followed by 1 pooling layer and 3 fully connected layers.
        \item HARCNN3 with 3 convolutional layers: The model consists of 3 convolutional layers, each followed by pooling layers, and 3 fully connected layers.
    \end{enumerate}
    \item \textbf{Architectures in HtFE$^{sen}$$_8$}: This model group builds on HtFE$^{sen}$$_5$ by further varying the stride configurations.
        \begin{enumerate}[left=0pt, labelsep=0.6em]
        \item In HARCNN1, the stride is varied to 1, 2, and 3.
        \item In HARCNN3, the stride is varied to 1 and 2.
        \end{enumerate} 
\end{itemize}

\subsection{The Tiny Auxiliary Model}
Since FML, FedKD, and FedMRL rely on a global auxiliary model for mutual distillation, it is crucial for this auxiliary model to be as compact as possible to reduce communication overhead during parameter transmission~\citep{wu2022communication}. Consequently, we select the smallest model within each heterogeneous model group to serve as the auxiliary model in all scenarios. 

\section{Additional Benchmark Results}

\subsection{Accuracy Curves in Text Modality}
\begin{figure}[htb]
    \centering
    \includegraphics[width=\linewidth]{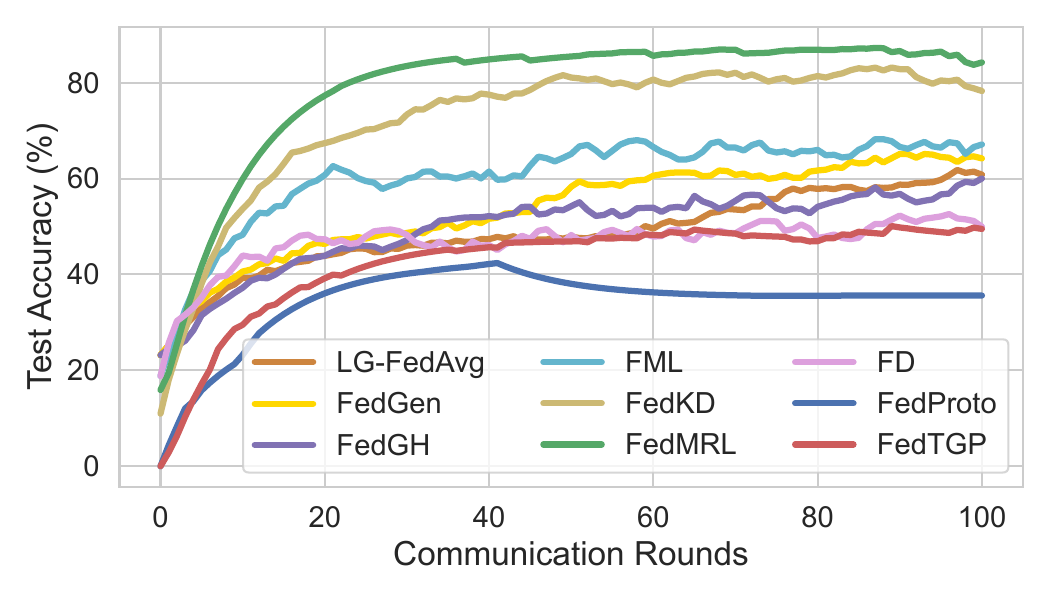}
    \caption{The test accuracy (smoothed) curves on the AG News dataset under Dirichlet settings using HtFE$^{txt}$$_6$.}
    \label{fig:test curve on text modality}
\end{figure}

In this part, we visualize the training curves of baselines on the AG News dataset under Dirichlet settings using HtFE$^{txt}$$_6$. As shown in Fig.~\ref{fig:test curve on text modality}:
(1) Mutual distillation demonstrates the fastest convergence and highest final accuracy, highlighting its robustness in scenarios with significant data and model heterogeneity. This advantage arises from the shared homogeneous auxiliary model, which remains less influenced by the variations across client models.
(2) Prototype sharing performs the worst among the three categories, showing slow convergence and low final accuracy. This underscores the challenge of obtaining effective prototypes in text modality tasks with strong model heterogeneity, limiting the overall effectiveness of prototype-sharing methods. 

\subsection{Accuracy Curves in Sensor Signal Modality}
\begin{figure}[ht]
	\centering
	\includegraphics[width=0.9\linewidth]{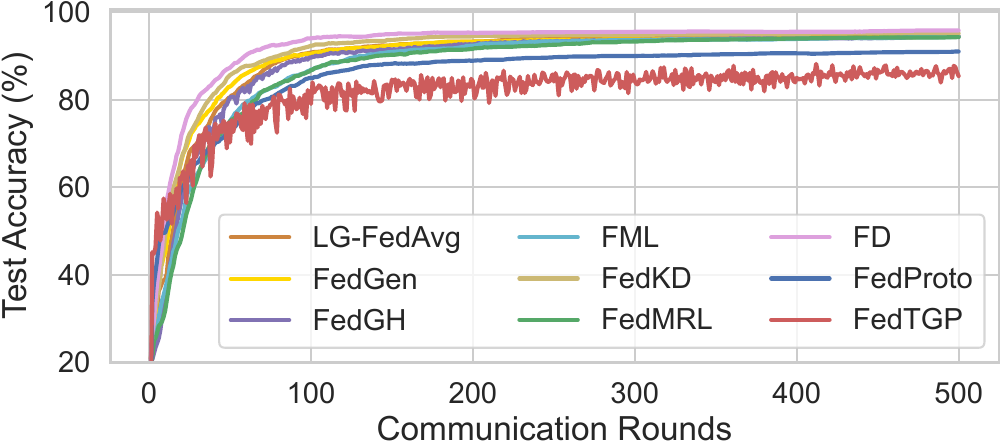}
	\caption{The test accuracy curves on the HAR dataset under real-world settings using HtFE$^{sen}$$_8$.}
	\label{fig:test curve HAR}
\end{figure}

\begin{figure}[ht]
	\centering
    \includegraphics[width=0.9\linewidth]{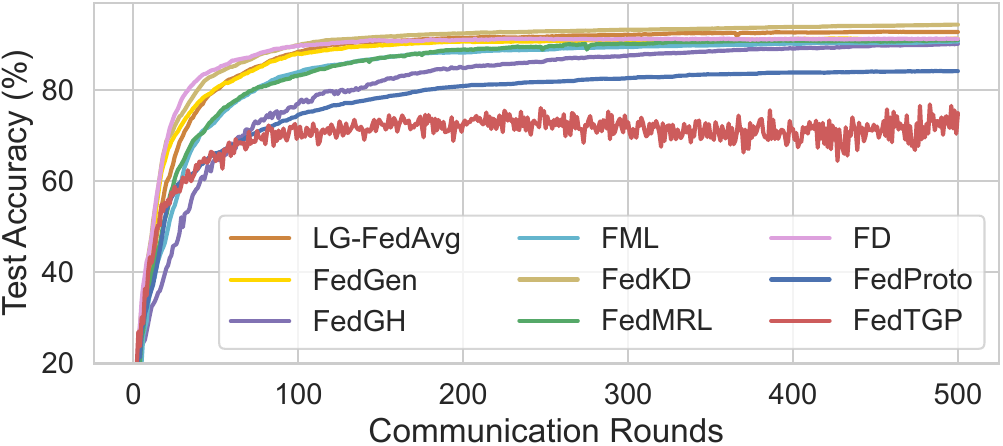}
	\caption{The test accuracy curves on the PAMAP2 dataset under real-world settings using HtFE$^{sen}$$_8$.}
	\label{fig:test curve PAMAP2}
\end{figure}

As shown in \cref{fig:test curve HAR} and \cref{fig:test curve PAMAP2}, we visualize the training curves of different methods on the HAR and PAMAP2 datasets under the real-world setting using HtFE$^{sen}$$_8$. From these curves, we know that: (1) Partial parameter sharing and mutual distillation methods exhibit smooth convergence, demonstrating their robustness to model heterogeneity. (2) In contrast, prototype-sharing methods like FedProto and FedTGP show poor performance, with FedTGP displaying particularly slow and unstable convergence. (3) Interestingly, FD converges quickly, highlighting the effectiveness of logits over prototypes for class representations in sensor signal modalities. This suggests that logits, which directly capture class probabilities, are more efficient for fast adaptation and decision-making, making them a promising direction for future research in sensor signal tasks.

\end{document}